\pdfoutput=1

\documentclass[11pt]{article}

\usepackage{acl}
\usepackage{enumitem}
\usepackage{times}
\usepackage{latexsym}
\usepackage{float}
\usepackage{subfig}
\usepackage{amsmath}
\usepackage{xcolor}  
\usepackage[font=small,labelfont=bf]{caption}
\usepackage{graphicx}
\usepackage{tabularx}
\usepackage{colortbl}

\usepackage[T1]{fontenc}

\usepackage[utf8]{inputenc}

\usepackage{microtype}

\usepackage{inconsolata}

\newenvironment{itemizesquish}{\begin{list}{\labelitemi}{\setlength{\itemsep}{-0.2em}\setlength{\labelwidth}{.5em}\setlength{\leftmargin}{\labelwidth}\addtolength{\leftmargin}{\labelsep}}}{\end{list}}

\title{A Monte Carlo Language Model Pipeline for Zero-Shot Sociopolitical Event Extraction}

\author{\quad Erica Cai \quad Brendan O'Connor\\
  University of Massachusetts Amherst \\
  \texttt{\{ecai,brenocon\}@cs.umass.edu} \\}

\begin{document}
\maketitle
\begin{abstract}
Current social science efforts
automatically populate event databases
of ``who did what to whom?''\ tuples,
by applying event extraction (EE) to text such as news.
The event databases are used to analyze sociopolitical dynamics between actor pairs (dyads) in, e.g., international relations.
While most EE methods heavily rely on rules or supervised learning,
\emph{zero-shot} event extraction could potentially allow
researchers to flexibly specify arbitrary event classes for new research questions.
Unfortunately, we find that current zero-shot EE methods, as well as a naive zero-shot approach
of simple generative language model (LM) prompting, perform poorly for dyadic event extraction;
most suffer
from word sense ambiguity, modality sensitivity, and computational inefficiency. 
We address these challenges with
a new fine-grained, multi-stage instruction-following generative LM pipeline,
proposing a Monte Carlo approach to deal with, and even take advantage of,
nondeterminism of generative outputs.
Our pipeline includes explicit stages of linguistic analysis (synonym generation, contextual disambiguation, argument realization, event modality), \textit{improving control and interpretability} compared to purely neural methods.  This method outperforms other zero-shot EE approaches, and outperforms naive applications of generative LMs by at least $17$ F1 percent points.
The pipeline's filtering mechanism greatly improves computational efficiency, allowing it to perform as few as 
12\%
of queries that a previous zero-shot method uses.
Finally, we demonstrate our pipeline's application to dyadic international relations analysis.
\end{abstract}

\section{Introduction}

\begin{figure}
    \centering
    \includegraphics[scale=0.26]{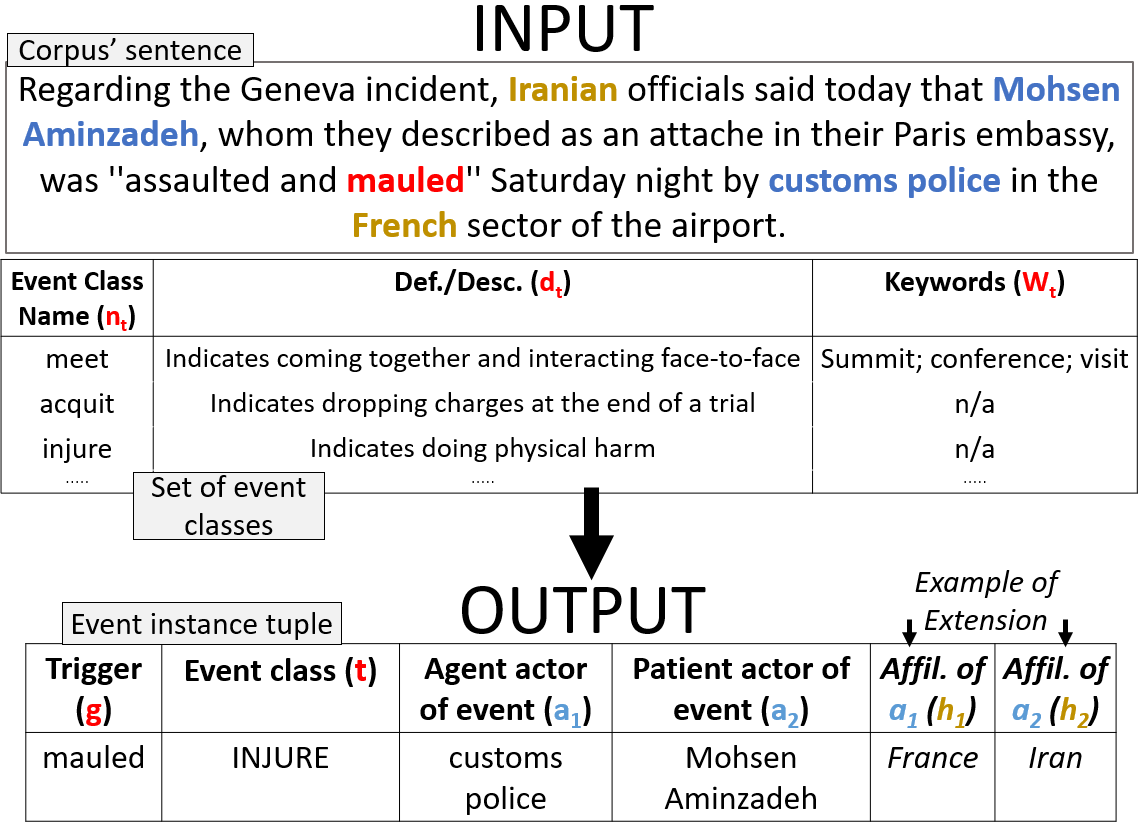}\vspace{-.8em}
    \caption{Example of zero-shot dyadic event extraction.
    Text from \emph{New York Times},
    July 14, 
    1987 
    \cite{SandhausNYT}. \vspace{-2em}}
    \label{fig:task}
\end{figure}

Event extraction (EE) infers actions and participants involved with them as structured \textit{events} from unstructured text data, and is useful for many areas such as knowledge graph construction and intelligent question answering \cite{gao2016collab,liu2020event, li2021document, cao2020deformer}. However, many EE methods require training on substantial, difficult-to-collect annotated data \cite{li2022survey,liu2016leveraging}.
\textit{Zero-shot} EE combats this challenge by providing flexibility to extract events from text without annotated data \cite{huang2018zero,zhang-etal-2021-zero,zhang-etal-2022-zero,zhang2022efficient,lyu2021zero,mehta-etal-2022-improving-zero,gao2016collab}, therefore  allowing users to ask for specific new types of events (e.g.\ cyberwarfare, environmental incidents) that may be enormously important for specific research needs, but which do not correspond to ones defined in previous event ontologies, where annotations may have taken years to produce (e.g.~in political science \cite{gerner2002conflict} or natural language processing \cite{doddington2004automatic}). 
The input for zero-shot EE is a corpus and a user-specified set of \textit{event classes} (types),
with textual names and optionally descriptions. The output are event \textit{instances} which contain structured details about each event class occurrence in text (Fig.~\ref{fig:task}).

\begin{figure*}
    \centering
    \includegraphics[scale=0.38]{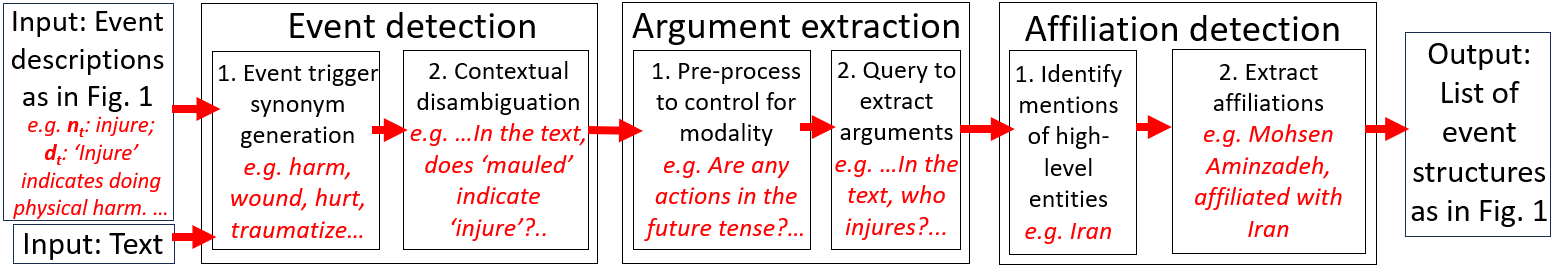}\vspace{-.7em}
    \caption{This work's multi-stage LM pipeline, where the event class for our running example from Fig.~\ref{fig:task} is \textsc{Injure}.
        \label{fig:pipeline}}
    \vspace{-1.60em}
\end{figure*}

\textit{Dyadic} EE extracts action occurrences between interacting pairs of actors (\emph{dyads}; e.g.\ \citet{wasserman_social_1994})  to construct graphs
that depict sociopolitical relations, where nodes correspond to actors and edges correspond to action occurrences. 
Earlier work has explored a variety of other methods for dyadic EE;
our view of the task follows original rule-based NLP efforts
in international relations \cite{Schrodt1994ValidityAO} 
that extract events between \emph{source} and \emph{target} actors
(i.e., linguistic theory's \emph{agent}-\emph{patient} roles \cite{dowty1991thematic,williams_2015}),
using pattern-based extraction from a fixed ontology \cite{Gerner1994,gerner2002conflict,schrodt2004event,Boschee2013AutomaticEO}.\footnote{While this line of work requires no annotated text, it relies on hand-engineering thousands of patterns.}
Machine learning approaches have also explored dyadic EE \cite{oconnor-etal-2013-learning, beieler-2016-generating, halterman2021extracting, Norris2017}, with applications in 
conflict prediction, social networks, and misinformation \cite{obrien2010crisis, brandt2011real, Smith2020AutomaticDO, stoehr2022ordinal}. In contrast to these knowledge engineering and machine learning approaches,
zero-shot EE promises to use significantly less expert knowledge and data,
aside from conventional pretraining outside the task.

Fig.~\ref{fig:task} illustrates components of dyadic EE on an example sentence.
Following prior (non-dyadic) EE NLP research, we divide it into subtasks
of
 \textit{event detection} to identify an event instance of a particular class (e.g.~\textsc{Injure}),
and \textit{argument extraction} to collect involved agent-patient actor pairs of people or organizations (\texttt{customs police}, \texttt{Mohsen Aminzadeh}).\footnote{More broadly, arguments of interest are actors with sociopolitically relevant agency; for example, military vehicles.}
Our main dyadic EE application also calls for
identifying higher-level entities, such as countries (\texttt{France} and \texttt{Iran}) that actors are affiliated with to analyze
their dynamics \cite{Schrodt1994ValidityAO}.

We seek to apply zero-shot EE to the dyadic EE task, but unfortunately find that many current zero-shot EE methods---based on, for example, textual entailment---suffer from word sense ambiguity, modality, and efficiency issues. For example, some methods only take an event class name from a user to define the event class, but this causes ambiguity
(e.g.\ is \emph{charge} about money, indictment, or attack?).\footnote{In the zero-shot setting, apparently only one work addresses this issue \cite{zhang2022efficient}.}
Methods may also be sensitive to event modality, i.e., whether the event instance really occurred, or is a hypothetical, or may occur in the future. Finally, some methods are very computationally expensive, because they require 
making inferences exhaustively for many text spans.

To address these challenges and improve performance, we propose using generative large language models (LLMs or simply LMs), which have significant potential to capture interesting aspects of meaning and nuances of language \cite{raffel2020exploring,brown2020language,rogers-etal-2020-primer,ouyang2022training,piantadosi2023modern}, but a simple application of them to zero-shot event detection produces poor results \cite{Gao2023ExploringTF}.
We further investigate by building on recent zero-shot approaches based on
text entailment (TE) \cite{lyu2021zero,yin2019benchmarking} to  develop naive generative LM baselines, but find poor performance.

Instead of naive querying, we propose using a generative LM in a \textit{fine-grained event extraction pipeline}, which allows for more explicit control of handling particular semantic properties.
This follows a very widespread approach in rule-based and non-neural EE, including early work \cite{gildea-jurafsky-2002-automatic,ahn2006stages,ji2008refining} which uses pipelines of classifiers and rule-based extractors to incrementally build and refine event structures, often building on ``classic NLP pipelines'' predicting linguistic structures like
 POS tags, parse trees, or named entities \cite{toutanova-etal-2005-joint,finkel-etal-2006-solving,bunescu-2008-learning,watanabe-etal-2008-pipeline,manning-etal-2014-stanford,peng-etal-2015-concrete}.
Instead of predicting categories from a general-purpose syntactic-semantic ontology, we query an LM for task-specific inferences. 

Fig.~\ref{fig:pipeline} illustrates our pipeline on the event class, \textsc{Injure} from Fig.~\ref{fig:task}. 
It implements event detection as lexical semantic stages of (1) event trigger synonym generation,
and (2) contextual disambiguation, to detect if a trigger usage indicates an event instance.
Next, argument extraction uses extractive QA 
\cite{du2020event,liu2020event} to identify arguments, and controls for event modality.
Finally,
we add detection of actors' affiliations with higher-level entities (e.g.\ countries, companies) 
to perform an international relations case study (\S\ref{s:affil_detect}).\footnote{ As a positive example, we note the CASE ACL-IJCNLP 2021 workshop \cite{case-2021-challenges}'s shared tasks on concrete sociopolitical event extraction tasks, such as fine-grained ACLED event detection.}

We propose a competitive zero-shot EE method that relies on generative LMs, with contributions:

\vspace{-.5em}
\begin{itemizesquish}
    \item \textbf{Monte Carlo} -- We propose probabilistically sampling many sets or single-value outputs from the LM at each stage, since LM outputs may be non-deterministic. This improves robustness, and through the temperature hyperparameter, allows the user to tune cost/performance tradeoffs in synonym generation (\S\ref{s:mc}).
    \item \textbf{Interpretability} --  Our pipeline improves interpretability compared to an all-at-once neural black box, allowing control for handling particular semantic properties.
    \item \textbf{Semantic challenges} -- Our approach addresses significant errors from lexical ambiguity and allows control for event modality.
    \item \textbf{Performance/efficiency} -- Our approach outperforms other recent zero-shot EE approaches on the widely-used Automatic Content Extraction (ACE) dataset \cite{doddington2004automatic,li2022survey}, outperforms a naive application of generative LMs by at least 17 F1 percent points, and could perform just 11.3\% of queries that a previous zero-shot TE approach performs. 
    \item  \textbf{Bringing EE beyond artifical NLP datasets} -- We demonstrate the flexibility of our approach for sociopolitical analysis. 

\end{itemizesquish} 
\vspace{-.6em}

\noindent We hope the flexibility and benefits of zero-shot generative LM pipelines can help support future semantic extraction applications.

\vspace{-.1em}
\section{The Zero-shot EE Task} \label{s:task}
\vspace{-.1em}

Formally, the input of our zero-shot EE task is the corpus' sentences $\mathcal{S}$
and event descriptions $\{\langle n_t,d_t,W_t\rangle \mid t \in \mathcal{T}\}$ where $\mathcal{T}$ is a set of event classes (e.g.~\textsc{Injure}), $n_t$ is an event class name (e.g.~{injure}), 
$d_t$ is a short definition or description of the event class, 
and $W_t$ is an optional set of keywords that help to describe it (possibly empty). Our task does not need annotated examples.

The output are sets of event instances for each sentence in $\mathcal{S}$ as in our running example in Fig.~\ref{fig:task}. Each event instance is a tuple $\langle t,g,a_1,a_2\rangle$, where:

\vspace{-.5em}
\begin{itemizesquish}
\itemsep-.3em 
\item $t\in\mathcal{T}$ is the \textit{event class} (e.g.~ \textsc{Injure}).
\item $g$ is the \textit{event trigger}, a word in the sentence that identifies 
the event class (e.g.~\texttt{mauled}).
\item $\{a_1, a_2\}$ are actor pair \textit{event arguments} explicitly mentioned in the sentence (e.g.~\texttt{customs police}, \texttt{Mohsen Aminzadeh}).
\end{itemizesquish}
\vspace{-.7em}

\noindent The terminology of triggers, arguments, and classes follows the EE literature.
We consider an additional variation to only extract actor participants as arguments---\emph{who did what to whom?}---%
where a \textit{dyadic} pair consists of one actor instigating the event (\textit{agent}, $a_1$) and the other receiving or being affected by it (\textit{patient}, $a_2$). Dyadic arguments---able to populate a graph with actor arguments as nodes and events between them as edges--- have tremendous social scientific utility to characterize social networks and relational dynamics, and
are grounded in \citet{dowty1991thematic}'s semantic proto-role theory.
Although dyadic EE may superficially seem similar to binary relation extraction, events mostly involve actions constrained to a specific time frame, therefore allowing analysis into how pair dynamics change over time; in contrast, relations tend to describe static associations \cite{agichtein2000snowball,yates2007textrunner,Cui2017ASO}.

Further, we extend our task to add \emph{affiliation detection}, extracting a higher level entity that each actor is affiliated with if any (see \S\ref{s:eval} and Fig.~\ref{fig:task}): 
\vspace{-.6em}
\begin{itemizesquish}
\itemsep-.2em 
    \item \textit{Additional input}, $\mathcal{C}$: The name of a higher level entity \textit{category} (e.g. country, company).
    \item \textit{Additional output}, $\langle h_1,h_2\rangle$ s.t. $a_1\in h_1$, $a_2\in h_2$, and $h_1,h_2 \in \mathcal{C}$ (possibly empty). (In Fig.~\ref{fig:task}, $h_1=$ \texttt{France}, $h_2=$ \texttt{Iran}, and $\mathcal{C}=$ country.)
\end{itemizesquish} 
\vspace{-.6em}
Such high-level entity actor information is useful for predicting future conflict or cooperation \cite{stoehr-etal-2021-classifying,brandt2011real}. 
For future reference, we refer to the subtasks of extracting outputs $\langle t,g \rangle$ as \textit{event detection}, $\langle a_1,a_2\rangle$ as \textit{argument extraction}, and $\langle h_1,h_2\rangle$ as \textit{affiliation detection}.

\section{Unreliability of Naive LLM Queries}  
\label{s:unrel}

\begin{figure*}
    \centering
    \includegraphics[width=1\textwidth]{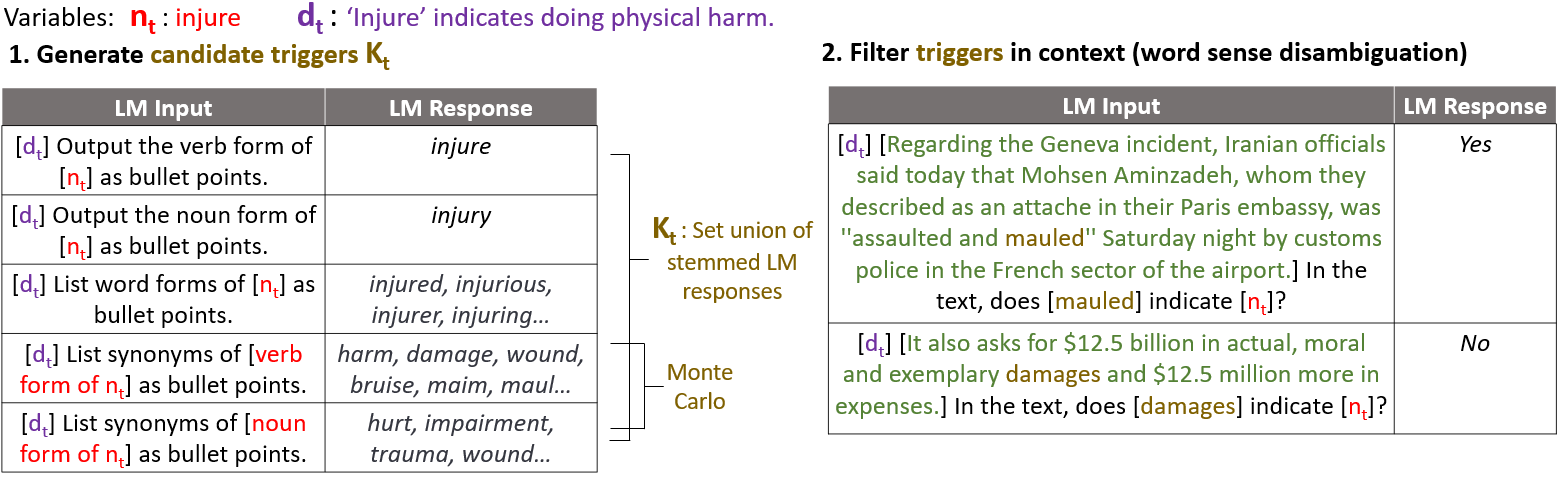}\vspace{-1em}
    \caption{Prompt-based pipeline for event detection (\S\ref{s:method_evt_detect}) on the running example from Fig.~\ref{fig:task}. \vspace{-1.8em}}
    \label{fig:evdect}
\end{figure*}

Given significant potential of generative LLMs in recent zero-shot work, we naively test them for event detection by transferring promising existing zero-shot TE or QA approaches to LLMs. Variations of a TE or QA-based approach have been studied in supervised EE \cite{du2020event,liu2020event}, zero-shot relation extraction \cite{levy-etal-2017-zero}, zero-shot text classification \cite{yin2019benchmarking}, and zero-shot EE \cite{lyu2021zero}. TE outputs a probability that a premise of text implies a hypothesis that the text contains an event instance (e.g.~\textit{This text is about injuring.}) \cite{lyu2021zero} (\citet{yin2019benchmarking} explore hypothesis phrasing).

Our naive approach is brute force, aiming to detect positive event instances by exhaustively performing TE over the Cartesian product of all sentences and hypotheses enumerating each event class. We use \textit{"This text is about..."} as the TE hypothesis (details in \S\ref{s:tcapp}), converting it to a boolean question to explore generative LLM performance (e.g.~\textit{Is the text about injuring?}) (row 1, Table 1). We add short definitions to prompts and hypotheses (rows 3,4) to address ambiguity and explore effects of wording, replacing \textit{"is about"} with \textit{"discusses"} (row 2) (\S\ref{s:tcapp}). Our evaluation is over every sentence in the same 40 documents and 33 event classes of ACE (\S\ref{s:eval}) as many EE evaluations and explores roberta and deberta (large) for TE and GPT3.5 (text-davinci-003) and ChatGPT (gpt-3.5-turbo) as generative LMs for QA.

\vspace{-.7em}
\begin{table}[H]
{\footnotesize
\begin{center}
\begin{tabularx}{\columnwidth}{ l |X X X X}
\hline
 &  \multicolumn{2}{|c|}{Text Entailment} & \multicolumn{2}{c}{Generative}\\\hline
 & Roberta & \multicolumn{1}{c|}{Deberta} & GPT3.5 & ChatGPT \\ 
 \hline\hline
 1 \citet{lyu2021zero} & 33.6 & \multicolumn{1}{c|}{31.1} & 37.6 & 40.7 \\ 
 2 Wording change & 22.4 & \multicolumn{1}{c|}{33.4} & 44.5 & 41.7 \\ 
 3 Add \textbf{def.} to [1] & 18.8 & \multicolumn{1}{c|}{4.3} & 35.9 & 29.0 \\ 
 4 Add \textbf{def.} to [2] & 21.8 & \multicolumn{1}{c|}{6.7} & 44.1 & 42.5  \\
\hline\hline
  
\end{tabularx}

\end{center}
}
\vspace{-.8em}
\caption{Simple query baseline F1 performance (classif.).}
\label{tab:promptvar}
\end{table}\vspace{-1em}


\noindent The results suggest that directly querying over long text spans may not be promising for improving performance. Table 1 shows some variation, which supports \citet{Gao2023ExploringTF} and \citet{yuan2023zeroshot}'s findings, and shows that adding a short definition in a TE hypothesis causes performance to plummet.

While our naive approach exhaustively queries on all pair combinations of sentences and hypotheses about event classes, queries are over entire sentences; it cannot detect multiple instances of the same event class in one sentence. Less naive approaches could instead query over substrings of sentences based on SRL model outputs, but this involves more computation and does not achieve significantly better performance \cite{lyu2021zero}.



\section{Event Extraction Pipeline with Generative Language Models}

\noindent
Given unreliable and poor performance of \textit{naive}, exhaustive generative LM prompting, we propose a pipelined instruction-following approach to LM-based, zero-shot dyadic event extraction, which includes (to our knowledge) the \textit{first competitive zero-shot event detection method using generative LMs}. It involves separate steps for event detection---%
with synonym generation, filtering, and disambiguation (\S\ref{s:method_evt_detect})---%
then argument extraction (\S\ref{s:method_arg_extract})
and later
affiliation detection (\S\ref{s:affil_detect}).
Throughout, it uses a Monte Carlo (MC) sampling method (proposed in \S\ref{s:mc}) to improve robustness,
and to control size and diversity of candidate trigger synonym sets.

\subsection{Event Detection}  \label{s:method_evt_detect}

\begin{figure*}
    \centering
    \includegraphics[width=1\textwidth]{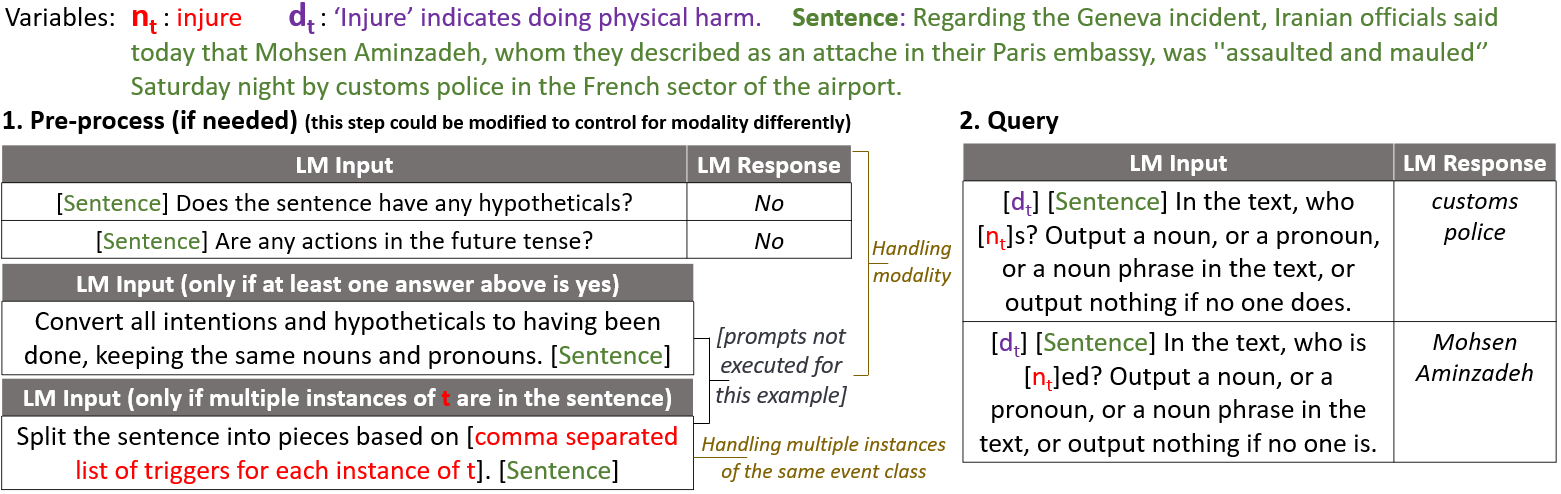}\vspace{-.7em}
    \caption{Prompt-based pipeline for argument extraction (\S\ref{s:method_arg_extract}) on the running example from Fig.~\ref{fig:task}.\vspace{-1.8em} }
    \label{fig:argex}
\end{figure*}

We propose a generative LM method for event detection which queries over words and phrases instead of over entire sentences, and illustrate the approach on our running example in Fig.~\ref{fig:evdect}. Our method is more selective about making queries than the naive exhaustive LM querying approach, and this greatly improves efficiency (\S\ref{s:eval}). 
Our method finds a trigger in the following phases:
\vspace{-.5em}
\begin{itemize}
\setlength{\itemindent}{0em}
\itemsep-.2em 
\item[\textit{\footnotesize Step 1.}] Generate a set of candidate trigger word stems $K_t$ for each event class $t$ given $\langle n_t,d_t,W_t\rangle$.
\item[\textit{\footnotesize Step 2.}] Identify if a candidate trigger stem $k_t \in K_t$, if in a sentence, is a stem of an actual trigger word for an event instance (disambiguation).
\end{itemize}
\vspace{-.4em}
The event detection output is sets of event type and trigger tuples $\langle t, g \rangle$ corresponding to each sentence. 

\vspace{.2em}
\noindent\textbf{Step 1: Generate candidate trigger terms.} \textit{For each event class $t$, generate a (possibly overcomplete) set of candidate trigger words and phrases, to identify event instances of the class, as $K_t$.}

$K_t$ is populated by expanding $n_t$ to many more lexical items, 
generating a set of its \textbf{inflections},
\textbf{noun} and \textbf{verb} forms of $n_t$, and their respective \textbf{synonym sets} (\ref{fig:evdect}).
This expansion is also performed for each $w_t \in W_t$; $K_t$ is the union of these expansions,
all generated by LM prompting
(Fig.~\ref{fig:evdect}; \S\ref{s:methapp}),
then stemmed \cite{porter2008snowball} and deduplicated. 
Each query includes definition $d_t$, helping the model use an appropriate word sense of $n_t$.
Synonym sets are generated with a Monte Carlo (MC) method (\S\ref{s:mc}).

For example, $n_t$=\emph{injure} yields 68 word stems in $K_t$, including near-synonyms (\emph{hurt}), many hyponyms (\emph{wound}, \emph{maim}), and some only moderately similar terms (\emph{torment}, \emph{loss}). We prefer to possibly overgenerate, since the next step removes spurious matches.
While we explore using an LLM for this lexical expansion, alternative resources such as word embeddings (e.g. GLOVE; \citet{pennington-etal-2014-glove}) or hand-built lexical databases (e.g. WordNet; \citet{miller1995wordnet}), could replace this step in future work.
Possible LLM advantages include accommodation of multiword $n_t$ and $w_t$ (e.g. \textit{start organization}), flexibly distant temperature-based MC control over synonym set size (\S\ref{s:mc}), and fast runtime---generating synonym sets takes only a few seconds. 

\vspace{.2em}
\noindent\textbf{Step 2: Filter triggers (disambiguation).} \textit{Within a sentence context, determine if candidate trigger stem $k_t$ actually identifies the event.}

Our method analyzes all sentences $s\in \mathcal{S}$ for any stems $k_t \in K_t$ of a class $t \in \mathcal{T}$.\footnote{Stems are matches as prefixes to words in $s$.}
Each match is disambiguated with a generative model, asking if the word that has $k_t$ as a substring indicates class $t$ (in the form of $n_t$ or $w_t$; step 2 of Fig.~\ref{fig:evdect}). Matches of the same stem are considered as different potential instances of the same event type.
The query includes definition $d_t$. 
A \textit{yes} answer indicates successful event detection, while \textit{no}, otherwise. Monte Carlo voting improves robustness (\S\ref{s:mc}).

\textbf{Summary.} This two-stage system (Fig.~\ref{fig:evdect}) helps ensure both efficiency and accuracy---Step 1 greatly reduces the number of sentences requiring LLM analysis compared to previous zero-shot EE approaches and the naive approach (efficiency analysis in \S\ref{s:mc}), while Step 2 protects against spurious matches.
We find that irrelevant trigger candidates from Step 1 do not change performance much (\S\ref{s:methapp}), but too many matches can significantly hurt disambiguation efficiency (see \S\ref{s:mc}).  Note that prompts and tense-insensitive stem matching naturally disregard modality, allowing detection of events with future, past, hypothetical, or other semantic modalities, which we could control for in \S\ref{s:method_arg_extract}.

\subsection{Event Argument Extraction}  \label{s:method_arg_extract}

To populate a graph depicting sociopolitical relations, we propose a multistage generative LM argument extraction method that is similar to that of \citet{du2020event} and \citet{liu2020event}, which use extractive QA to determine arguments (e.g.~\textit{“Who injures?”}, illustrated in Fig.~\ref{fig:argex}, following the running example in Fig.~\ref{fig:task}). For each input tuple of event type and trigger ($\langle t,g\rangle$) from event detection, a corresponding actor argument tuple $\langle a_1,a_2 \rangle$ (if existing) is output. Our task aims to extract actor pairs only, not single arguments.

\vspace{.2em}
\noindent\textbf{Query step.} Querying extracts dyadic agent and patient actors $\langle a_1, a_2 \rangle$ for each event instance. Each query is over a sentence or a modified output of \textit{pre-processing}. For event names that are regular verbs, the query  has the form \textit{Who [$n_t$]s?}, \textit{Who is [$n_t$]ed?} (Fig.~\ref{fig:argex}); this format has simple modifications for accommodating other forms of $n_t$ (e.g.~multi-words) and ensuring grammatical correctness 
 (\S\ref{s:methapp}). The query includes definition $d_t$ (addressing ambiguity) and an instruction to specify that the answer belongs to the text. MC (\S\ref{s:mc}) increases robustness.

\vspace{.2em}
\noindent\textbf{Pre-process step.} Pre-processing text (if needed) before querying allows \textit{controlling for modality:} since query phrasing assumes a current or past-tense event instance, the method asks (1) if a sentence has hypotheticals or intentions; if so, it (2) converts them to past-tense using an LLM instruction (Fig.~\ref{fig:argex}). The step could easily be modified to consider only past-tense events (remove pre-processing) or consider additionally just future-tense events (remove consideration of hypotheticals), depending on the application. In addition, to extract arguments for each unique instance of the same event class in a sentence (if more than one), the method prompts to split the sentence into text spans corresponding to each instance of $t$ (Fig.~\ref{fig:argex}).

\section{Monte Carlo (MC) for Synonym Set Generation and Robustness}  \label{s:mc}

We propose an MC approach that is useful for generating lexical resources and producing more robust outputs, positively impacting performance and efficiency of our pipeline. Generative LMs are nondeterministic, which benefits our approach by allowing construction of diverse cumulative synonym sets from many samples of synonym sets, but decreases output robustness on QA tasks. 

\vspace{.2em}
\noindent\textbf{MC to generate synonym sets.} MC allows control for broadness of a synonym set while balancing compute cost during event detection. Let a word's synonym set be $A$: our method populates it by repeatedly prompting to create synonym sets $A_y$: 
\vspace{-.5em}
\begin{equation} \label{eq2}
\begin{split}
A &= \cup_{y=1}^Y A_y
\end{split}
\end{equation}
\vspace{-1.9em}

\noindent i.e.~any $a\in A$ must appear in at least one sample.\footnote{Future work could explore higher thresholds of $\tau$, giving another method of controlling $A$'s breadth.} Our goal is to construct broad synonym sets (or over-generate --- while irrelevant synonyms in a trigger stem set do not affect performance much
(\S\ref{s:methapp}), they decrease efficiency, necessitating more queries during filtering (disambiguation step)).

\vspace{-.9em}
\begin{figure}[H]
    \centering
    \includegraphics[scale=0.19]{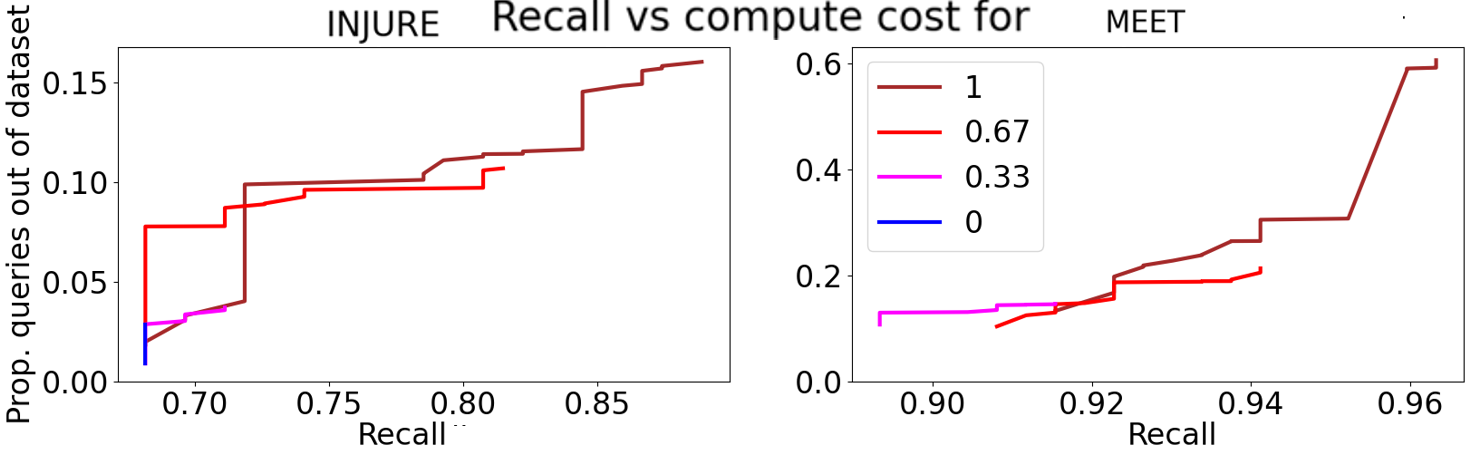}\vspace{-.7em}
    \caption{Recall vs.\ compute cost for \textsc{Injure} (left) and \textsc{Meet} (right) in ACE for different temperatures.}
    \vspace{-1.2em}
    \label{fig:reccost}
\end{figure}

\textit{Recall and efficiency.} Fig.~\ref{fig:reccost} shows that increasing the temperature hyperparameter, which controls the extent of LM output randomness, corresponds with higher recall ($\%$ of event instances for $t$ that have triggers in $K_t$) but also higher compute cost ($\%$ of queries performed out of number of sentences in ACE) as a tradeoff. Further, temperature $1$ may result in far lower efficiency. Yet, the efficiencies are significant improvements to TE (\S\ref{s:eval}).

\vspace{-.9em}
\begin{figure}[H]
    \centering
    \includegraphics[scale=0.2]{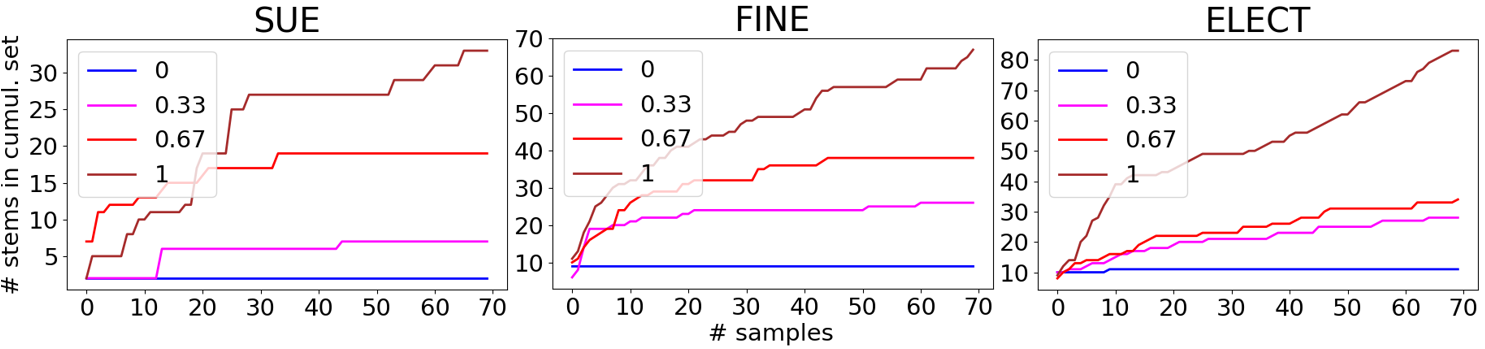}\vspace{-.7em}
    \caption{Cumulative set sizes over $70$ samples for 3 events of temps. in range $\{0,0.33,0.67,1\}$. \vspace{-1.5em}}
    \label{fig:cumset}
\end{figure}

Additionally, higher temperatures correlate with larger cumulative synonym sets, which nearly converge for temperatures $\leq 0.67$ (e.g.~\textsc{sue}, \textsc{fine} in Fig.~\ref{fig:cumset} converge for temperature $0.67$.). Therefore, drawing more samples at lower temperatures does not seem likely to increase synonym set sizes much.

In summary, temperature is crucial for controlling synonym set size (Fig.~\ref{fig:cumset}) which corresponds to increasing recall (Fig.~\ref{fig:reccost}), but may also incur higher compute cost. Informed by these results which also hold over the rest of the event names in ACE, our MC approach generates synonym sets with temperature $0.67$ over $70$ samples. While an alternative of MC is to run a single LLM prompt to generate a specific number of synonyms, this is not flexible: in Fig.~\ref{fig:cumset}, after 70 samples for temperature $1$, \textsc{elect} has 80 unique synonyms but \textsc{sue} has 30.

\vspace{.2em}

\noindent\textbf{MC for pipeline QA tasks.} Unfortunately, even for temperature $0$, LM outputs (GPT-family) are non-deterministic, which hurts robustness of our pipeline's QA outputs. We experiment and observe that most answers out of $10$ are unanimous, and if any, only a few (typically one) differ -- the proportion of unanimous boolean outputs using our system on GPT3.5 is $0.9986$, $0.9938$, and $0.9887$ for temperatures $0$, $0.33$, and $0.67$ respectively.
Our method takes the majority (similar to self-consistency prompting; \citet{wang2023selfconsistency}) from $9$ samples of each query in \S\ref{s:eval} but could sample much fewer (e.g.~$3$; see \S\ref{s:mcapp}). 

\vspace{.2em}
\noindent \textbf{Hyperparameter and prompt sensitivity.} Our fine-grained pipeline approach is less susceptible to hyperparameter and prompt variability: precision of event detection does not decrease with temperature because the disambiguation step tends to prevent irrelevant synonyms from counting towards false positives (\S\ref{s:methapp}). Number of samples does not need to be tuned; we advocate a large number for generating cumulative synonym sets, which nearly converge at large sample sizes (Fig.~\ref{fig:cumset}), and small for QA, where it has little effect. 
Finally, since tasks are fine-grained, we were able to choose the first prompts that we tried on the validation set without prompt engineering (\S\ref{s:methapp}), and MC helps to further weaken variability.

\section{Intrinsic} Evaluation \label{s:eval}

We first perform an intrinsic evaluation of event detection and argument extraction on the Automatic Content Extraction (ACE) dataset \cite{ahn2006stages}, which contains event class, trigger, and argument annotations over 33 event classes and 598 documents of news articles, conversations, and blogs. The ACE dataset is known as most popular for evaluating EE; many evaluations use the same train, validation, and test split \cite{ji2008refining,liao-grishman-2011-acquiring,hong2011using,nguyen2015event,nguyen2016joint-event,liu2016leveraging,huang2017liberal,sha2016rbpb,chen2017automatically,huang2018zero,liu2017exploiting,zhang2019joint,wadden2019entity,chen2020reading,du2020event,liu2020event,ahmad2021gate}. Our evaluation is over the same 40 documents as many EE evaluations, following \citet{wadden2019entity}'s pre-processing and modifications  \cite{cai-oconnor-2023-evaluating}. We verified LM instructions on the same 30 validation documents used by others (\S\ref{s:limitations}). 

In Table 2, we present results of our pipeline using GPT3.5 \cite{ouyang2022training}\footnote{We performed all experiments for this paper between Apr. $20$ and May $18$, using \texttt{text-davinci-003} and \texttt{gpt-3.5-turbo}}, which we replicated over $2$ runs. Unfortunately, baselines performing both event detection and argument extraction tasks in zero-shot EE are sparse. 

\textbf{Event detection.} We provide F1 micro-average performance of event detection (alternatively, \textit{trigger classification}; \S\ref{s:aceapp}) for our pipeline and baselines in column 3 of Table 2, where a true positive corresponds to a correctly identified event class ($t$) in a sentence. Our approach outperforms the other recent zero-shot EE approaches (see \S\ref{s:aceapp} for populating $d_t$, $W_t$, and complications in ACE). Its performance also exceeds that of our best-performing naive exhaustive LM query baseline (\S\ref{s:unrel}) by 16.7 F1 percent points, and is much closer to that of supervised EE than the other zero-shot EE performances are.

\vspace{-1em}
\begin{table}[H]
{\scriptsize
\begin{center}
\begin{tabular}{ c c c c c}
 Setting & System & Ev Dect  & Arg Ext & Arg Ext \\
 &  &   & (non-dyad)  &  (dyad)\\[0.5ex] 
 \hline\hline
 supervised & Lin et al 20 & 74.7  & 56.8 & -\\\hline
 zero-shot & Huang et al 18 & 49.1  & 15.8 & -\\
  &  Zhang et al 21 & 53.5  & 6.3 & -\\ 
  & Lyu et al 21 & 41.7 & 16.8 & 22.4 \\ 
  & \emph{This work} & \textbf{61.2} & n/a & \textbf{28.6} \\  
\hline
 zero-shot & Naive exhaust (\S\ref{s:unrel}) & 44.5  & n/a & n/a\\ \hline

\end{tabular}

\end{center}
\vspace{-.9em}
\caption{F1 micro-average performance of our and other methods on ACE, on event detection and argument extraction. }}\vspace{-1.2em}
\end{table}

\textbf{Argument extraction.} For reference, column 4 shows non-dyadic argument extraction performance (not applicable for our method) on ACE, where each event class corresponds to argument types beyond actors (e.g.~\textit{location}), and F1 performance is calculated based on correctness of individual arguments---the performances seem poor.

Finally, we provide dyadic argument extraction performance on ACE, counting a true positive only when the event and \textit{both} actors ($t$,$a_1$,$a_2$) are correct. For each event class in ACE, we map pairs of argument types (e.g.~\textit{\{Agent,Victim\}}; \textit{\{Attacker, Target\}}; \textit{\{Prosecutor, Defendant\}}) to analogous \textit{agent} and \textit{patient} types defined in \S\ref{s:task} (if they exist) to serve as $a_1$ and $a_2$. Evaluation is over the subset of event classes (20 out of 33) that have mappable dyadic arguments (\S\ref{s:aceapp}). Since this zero-shot subtask is new, we aim to adapt other evaluations to this setting. However, we find incompatibility between code for the zero-shot baselines and models that they depend on (e.g.~SRL), which underwent extensive updates. We thankfully got access to outputs of \citet{lyu2021zero}'s approach, and reproduced our dyadic evaluation setting for it to serve as a competitive baseline. Our dyadic approach outperforms it (col. 5), which we find outperforms other methods for non-dyadic argument extraction (column 4).



\textbf{Alternatives.} Besides GPT3.5 (used for the Table 2 evaluation), we could use ChatGPT, which is much cheaper and outperforms the baselines (e.g.\ 58 F1 on event detection). Using open-source generative LMs would have research and application advantages, but we find Llama 2 70B \cite{touvron2023llama} and Alpaca fail to produce diverse synonym sets and inconsistently produce inadequately formatted outputs for synonym generation (\S\ref{s:limitations}). Without diverse synonyms, the recall of our method, which only detects events upon trigger stem matches, is poor and therefore F1 performance is poor. We hope future work could explore more open-source LMs, which are rapidly improving.

\textbf{Efficiency analysis}. Our approach is efficient because it uses candidate triggers as a filter for performing LM queries. To analyze efficiency gains, we count boolean LM queries for disambiguation (\S\ref{s:method_evt_detect}), which dominates pipeline runtime (by contrast, argument extraction queries are sparse and synonym set generation queries do not vary with the size of the data). After considering MC, we find our method performs between 3\% and 30\% of \citet{lyu2021zero}'s queries, varying due to selecting between 1 to 9 MC samples per query, where a small number (e.g.~3, leading to performing 11.3\% of \citet{lyu2021zero}'s queries) seems sufficient (\S\ref{s:mc}).

\section{Affiliations and International Relations}  \label{s:affil_detect}

Next, we demonstrate the utility of our method on a real use case. A benefit of our dyadic EE pipeline is flexibility to add new application-specific components.  To define vertexes in the resulting dyadic event graph, we extend the pipeline with country-level affiliation detection for international relations analysis.

Specifically, our extension has a sociopolitical application of extracting events between high-level entities that actor arguments are representatives of (e.g.~countries; companies; any type of organization). The task input (\S\ref{s:task}) are sets of event instance tuples of type, trigger and arguments $\langle t,g,a_1,a_2\rangle$ from argument extraction, and $\mathcal{C}$, which is a higher level entity \textit{category}. Output are event instances with additionally affiliations of arguments $\langle h_1,h_2\rangle$ where $h_1,h_2 \in \mathcal{C}$ (in Fig.~\ref{fig:task}, $h_1$ and $h_2$ are \texttt{France}, \texttt{Iran}). In the case study, we set $\mathcal{C}$ as \textit{countries} and \textit{rebel groups}. To perform affiliation detection, our approach (1) finds all mentions of countries, and (2) determines if $a_1$ or $a_2$ is affiliated with any.

\vspace{.2em}
\noindent
\textbf{Step 1: Find country references.} Our method finds country references through keyword matching, primarily using the \texttt{CountryInfo.txt} database \citep{DVN-NBPRDW-2015}\footnote{\scriptsize\url{https://github.com/openeventdata/CountryInfo}}\footnote{\scriptsize \url{https://github.com/brendano/OConnor_IREvents_ACL2013}}
which includes noun, adjective, acronym, and misspelled references to countries (\S\ref{s:affapp}).
To associate cities or towns to a particular country, our method uses SpaCy 3 to identify all \texttt{GPE} and \texttt{NORP} named entities and checks if any is a location in a country by using the geocoding \texttt{Nominatim} API (having access to \texttt{OpenStreetMap} data).\footnote{\scriptsize\url{https://www.openstreetmap.org/}}
It selects the top country output if a named entity is ambiguous (toponym disambiguation). To evaluate, we select $100$ sentences from the New York Times Corpus (LDC2008T19 \cite{SandhausNYT})
that mention at least two countries based on the dictionary, and compare our annotations against this step's, finding $100\%$ accuracy. 

\vspace{.2em}
\noindent\textbf{Step 2: Extract affiliation.} To determine if an actor argument extracted by dyadic EE is affiliated with a country, our method (1) identifies affiliation if any country mention is directly part of the argument span ($a_1$ or $a_2$) or (2) iterates through each country mention and asks an LLM, \textit{"In the text, is [actor] affiliated with [country mention]?"}, applying MC to increase robustness. To evaluate, we select 100 samples that our method identified actors and countries for, and compare our annotations against the method, finding 86\% accuracy.

\vspace{.2em}
\noindent\textbf{Data and results.} We demonstrate our method and extension on NYT articles in 1987-1988 over self-defined event classes, using ChatGPT for event detection (10x cheaper than GPT3.5 with similar performance (\S\ref{s:eval})) and GPT3.5 for the other steps.

\vspace{-1em}
\begin{figure}[H]
    \centering
    \includegraphics[scale=0.2]{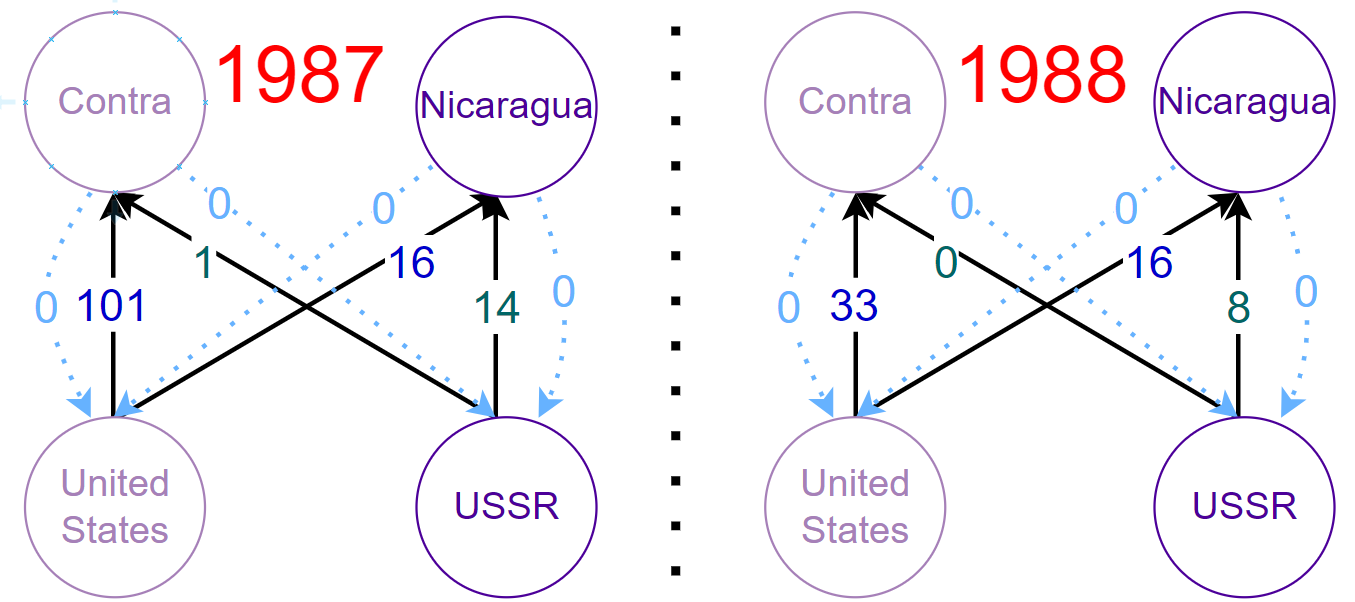}\vspace{-.7em}
    \caption{Dyadic event frequency network of \textsc{aid} in the proxy war during 1987 and 1988.}\vspace{-1.2em}
    \label{fig:aid}
\end{figure}

\textit{Proxy war.} In Fig.~\ref{fig:aid}, we observe interactions that are consistent with events during the 1980s, where US-backed rebels, the Contras, fought against the USSR-backed Nicaraguan government. This was a major Cold War proxy conflict. In 1987 and 1988, most of the USSR's aid went to Nicaragua, rather than the Contras ($93\%$, $100\%$ in $1987$, $1988$ respectively) and vice versa regarding the US's aid. In 1988, the US seems to aid Nicaragua slightly more because NYT began referring to Contras as the ``new Nicaraguan government.'' 
We also find neither Contras nor Nicaragua aids the US or USSR, as a simple illustration of  center-periphery imperial dependency structure \cite{Galtung1971}.

\section{Event Extraction Literature}

While our approach is zero-shot, annotated event instances have been critical for most EE, from older pattern matching approaches, for helping to learn and apply rules for identifying instances and arguments from surrounding context \cite{riloff1993automatically,riloff1999learning,califf2003bottom,liao-grishman-2011-acquiring,Boschee2013AutomaticEO}, to feature-based approaches, for helping to learn effective features for statistical models \cite{ji2008refining,gupta2009predicting,hong2011using,li2013joint,mcclosky2011event}, to deep learning approaches \cite{zhang2019joint,nguyen2019one,li2020event,lin2020joint,li2021document,ahmad2021gate}. For methods that do not use annotations \cite{huang2018zero,mehta-etal-2022-improving-zero}, a common approach is to learn from text that contains seed words to identify event classes  \cite{riloff1999learning,yangarber2000automatic,yu-etal-2022-building}. \citet{zhang-etal-2021-zero} present a zero-shot EE method that generates cluster embedding representations for event classes from sentences that contain synonyms of event names, and computes cosine distance from event clusters to classify instances. \citet{zhang2022efficient} refine this, addressing ambiguity by using event definitions to form these event cluster representations. 

The ambiguity challenge is also well-known in non-zero-shot EE, where \citet{liao2010filtered} and \citet{liao-grishman-2011-acquiring} design methods to ``limit the effect of ambiguous patterns", \citet{ji2008refining} try to learn ``correct" word senses during training, and \citet{liao2010using} use information about other event classes to resolve ambiguities in given instances.

\section{Conclusion}

We propose a multi-stage, fine-grained, generative LM pipeline for dyadic zero-shot EE that addresses ambiguity and efficiency challenges, controls for modality, allows for more interpretability, and outperforms other zero-shot EE methods.  Our pipeline benefits from an MC approach, which future work could explore and apply on other tasks (e.g.,~involving lexical resource generation). As zero-shot, the pipeline is deployable in various real-world settings and we demonstrated it on a case study in international relations, an important application.

\section{Limitations}
\label{s:limitations}

Our pipeline introduces a perspective of performing EE in fine-grained tasks and for implementation, we make choices about certain models, hyperparameters, and prompts.

\textbf{Models.} The Table 2 pipeline results use GPT3.5 (text-davinci-003), and we find that the pipeline using ChatGPT (gpt-3.5-turbo) also outperforms the zero-shot baselines. Further, these models outperform open-source generative models on specific tasks (\S\ref{s:eval} and \S\ref{s:modelsapp}), despite limitations of being proprietary and non-deterministic (our pipeline addresses model non-determinism in \S\ref{s:mc}). 
While we tried open-source models including 
Alpaca\footnote{\url{https://github.com/replicate/cog_stanford_alpaca}} and Llama (1 and 2) 7B, 13B, and 70B, Alpaca produced unreliably formatted outputs, and Llama 2 struggles to construct diverse synonym sets that are relevant to a word (e.g.~\textit{help, aid} as synonyms of \textit{injure}). However, such models may potentially be valuable for the future --- improving zero-shot performance of open-source models is an active research area.

\textbf{Hyperparameters.} The temperature hyperparameter is broadly an issue in many generative approaches, and we consider its effect on various factors in \S\ref{s:mc}. 

\textbf{Prompts.} We are very concerned about prompt sensitivity which we discuss in \S\ref{s:mc}; at a high level, our approach  has well-defined subtasks where small prompt changes do not impact results much (in contrast to a black-box neural net approach). 

MC helps to weaken variability in our approach. For example, our final synonym selection prompt was: \textit{List synonyms of [text] in bullet points.} Yet, alternative phrasings exist, such as: \textit{Output synonyms of [text] in bullet points.}, and \textit{What are the synonyms of [text]? Answer in bullet points.}, and \textit{List synonyms of [text] as a numbered list.} While these alternatives may construct a slightly different synonym set after a single sample, the variance among the synonym sets after 70 draws on any of these prompts at a given temperature is very similar to that of the synonym sets after 70 MC draws on our final prompt. We chose the final prompt out of convenience; it has fewer tokens and bullet point outputs are a simple format to parse.

We selected the first prompt that we tried for the rest of the tasks using the validation set discussed in \S\ref{s:eval}. While future study could explore different phrasings, our decision process suggests that with explicit and specific linguistic task definitions for word sense disambiguation, event argument extraction, and other tasks, prompt engineering may not be necessary for producing a higher accuracy output. We also find that minor changes to a prompt such as using "sentence" versus "text" and reordering the prompt does not affect the output. 

The one task that we did not use LLMs for involves identifying country references in a sentence, which we found is vulnerable to bias (\S\ref{s:risksapp}). We also found many errors in the output when trying LLMs and therefore opted for using a dictionary.

\section*{Version history}
An earlier version of this work was posted as arXiv:2305.15051, Cai and O'Connor (v1; May 2023),
with a later version presented at the NeurIPS Instruction Tuning and Instruction Following Workshop, December 2023.
This current version (arXiv:2305.15051 v2) incorporates revisions to the writing and presentation, and no changes to experimental results.

\section*{Acknowledgments}
We thank the UMass NLP group and anonymous reviewers for feedback.
Thanks also to Qing Lyu for help in analyzing their prior experiments.
This work was supported by NSF CAREER 1845576.
Any opinions, findings, and conclusions or recommendations expressed in this material are those of
the authors and do not necessarily reflect the views
of the National Science Foundation.

\bibliography{custom}

\appendix

\section{Risks}
\label{s:risksapp}

Large generative models may produce harmful word sequences, but our fine-grained pipeline avoids tasks that are more likely to produce open-ended answers, focusing on tasks that produce text with constrained output (e.g.~boolean QA, extractive QA, and synonym set generation). The one task that could face more bias from an LLM involves extracting a country mention from text in the affiliation detection extension, where a generative LM could assert what counts and does not count as a country, which could be controversial (e.g. Palestine). However, we do not use a generative model for this step because we observed more errors in addition to this potential bias.

One important application of our approach, which is extracting sociopolitical events, may be of great interest to social scientists (e.g.\ CASE workshop; \cite{nik-etal-2022-1cademy,dai-etal-2022-political,you-etal-2022-eventgraph}) as well as having government and military intelligence utility (see ethical discussion of \cite{li-etal-2020-gaia} on dual use issues for their multimodal tracking and surveillance system).

\section{Models}
\label{s:modelsapp}

For the results in Table 2, we use GPT3.5, but ChatGPT performs similarly and also outperforms the other zero-shot EE methods. For the demonstration of our method with the affiliation detection extension, we use ChatGPT to save cost.

We also try implementing our approach with open-source models which are useful for keeping data local and for having a reliable tool to implement the pipeline with. Further, some open-source models are more replicable; for example, Llama and Alpaca will consistently output the same response when the temperature is $0$. We try Llama (1 and 2) 7b and 13b, and Alpaca 7b (instruction-tuned version of Llama 1). Although Llama 2 is able to follow instructions (where Llama 1 struggles) and answer tricker questions such as: \textit{...it hurt their chances.$\backslash$n In the sentence, does `hurt' indicate `injure'?} (where Alpaca struggles), all inconsistently struggle with producing outputs of a specified format and with generating diverse synonym sets that are consistent with a word’s definition. For example, Llama 2 outputs \textit{help, assist, aid}, and \textit{support} as synonyms of \textit{injure}. Yet, Alpaca and Llama 2 serve as positive steps toward improving open-source models.

For the future, we foresee that not only will generative GPT-family LMs improve their performance on the fine-grained tasks in our pipeline, which will contribute to further performance gains, but also that higher performance open-source models will be available for use, which may help to increase replicability of results and eliminate cost issues for the using GPT-family models. We encourage future work to explore more open-source LMs for the pipeline.

\section{Naive Event Detection Details}
\label{s:tcapp}

For the naive generative LM event detection experiments in \S\ref{s:unrel}, we present more details about the experiments, including examples for prompts and further approaches that we tried. 

\textbf{Example of hypotheses.} An example of the four hypotheses described in Table 1, where definitions come from or have small modifications of  descriptions in the ACE annotation guidelines, is: 
    \begin{table}[H]
\small
\centering
\begin{tabular}{|l|l|}
\hline
\textbf{Hypotheses}  \\
\hline
1 This text is about pardoning. \\
\hline
2 This text discusses pardoning.\\\hline
3 This text is about pardoning, where `pardon' is  to lift \\a sentence imposed by the judiciary.\\\hline
4 This text discusses pardoning, where `pardon' is  to lift \\a sentence imposed by the judiciary.\\\hline

\end{tabular}
\caption{Example of hypotheses used for Table 1.} 
\end{table}
We also try other types of hypotheses, such as using root words of event class names instead of a more natural form of the event class name with an affix of -ing. 
    \begin{table}[H]
\small
\centering
\begin{tabular}{|l|l|}
\hline
\textbf{Hypotheses}  \\
\hline
 This text is about `pardon'. \\
\hline
 This text discusses `pardon'.\\\hline
 This text is about `pardon', where `pardon' is  to lift \\a sentence imposed by the judiciary.\\\hline
 This text discusses `pardon', where `pardon' is  to lift \\a sentence imposed by the judiciary.\\\hline

\end{tabular}
\caption{Example of alternative hypotheses.} 
 
\end{table}

\textbf{Converting hypotheses to boolean queries.} To convert hypotheses into boolean queries, for the hypotheses that include `about', the queries begins with:

`\textit{Is the text about...}'

\noindent For hypotheses that include `discuss', the query begins with:

`\textit{Does this text discuss...}'

\textbf{Model details.} We experimented with the boolean queries when using GPT, ChatGPT, Alpaca, and the roberta model fine-tuned on the BoolQA dataset. However, we found very poor performance using the roberta model fine-tuned on BoolQA, supporting \cite{lyu2021zero}'s finding.

For the text entailment experiments, we selected results using the probability threshold of 0.45 for deberta and of 0.55 for roberta since those thresholds tended to yield higher F1 performance scores over different variations of prompts. 

\textbf{Further alternative prompt strategies and models.} We also tried experiments on other hypothesis and prompt variations and tried other models. The experiments on deberta-xlarge and bart-large models yielded similar performances as those in the table. The prompt "\textit{Someone was pardoned}" referred to in  \cite{lyu2021zero} also did not produce any significantly different performance result. For definitions, we tried placing them before and after the "\textit{This text is about...}" hypothesis or its query counterpart in generative LMs. Further, we experimented with longer and shorter definitions, finding that the performance of these variations was similar to the performance of the variations in Table 1. 


\section{Event Detection and Argument Extraction Details}
\label{s:methapp}

Our event detection step allows for any single-word or multi-word event name $n_t$. However, while event detection works for any single- or multi-word event name, argument extraction requires the event class to have the potential to contain dyadic actor arguments -- for example, \textsc{injure} could have dyadic actor arguments because an agent actor $a_1$ could cause the \textsc{injure} action and a patient actor $a_2$ could be the receiver of the \textsc{injure} action. On the other hand, the verb event class \textsc{stand} may not be able to have dyadic actor arguments because an agent actor $a_1$ could instigate the instance, but no patient actor $a_2$ exists. Our pipeline assumes that each event class in the input could possibly have dyadic actor arguments.

The queries in argument extraction use the verb form of event name $n_t$. If $n_{verb,t}$ is a regular verb, the queries for extracting dyadic actors $a_1,a_2$ are straightforward as \textit{Who [$n_{verb,t}$]s?} and \textit{Who is [$n_{verb,t}$]ed?} (e.g. \textit{Who injures?} and \textit{Who is injured?}). When $n_{verb,t}$ is more complicated, containing a preposition as in \textsc{protest against}, argument extraction asks questions in the form of \textit{Who [$n_{verb,t}$]s?}  and \textit{Who is [$n_{verb,t}$]ed [preposition]?}. For example, the query for \textsc{protest against} is \textit{Who protests?} and \textit{Who is protested against?}. From an error analysis of our experiments, we find that our system achieves the highest performance on event classes that are single regular verbs, but could still correctly detect and extract arguments for multi-word event classes. 

\textbf{Extra Candidate Triggers.} From performing experiments to vary the size of each candidate trigger set and observing how many queries our method performs given a particular $K_t$ (as in the recall vs compute cost plot (Fig 7) of Sec 6), we found that including extra and irrelevant candidate triggers in the candidate trigger set leads to higher compute cost, but performance does not change much.

\textbf{Not susceptible to grammar changes or errors.} While our pipeline aims to construct questions that are grammatically correct, even if it makes a mistake in the affix of a word or in using the correct article, we find that generative models are able to answer questions correctly. This is different from text entailment models where performance probabilities may be more sensitive to small grammatical errors, depending on the model.

\textbf{Fewer alternatives for prompt wordings.} Given the specific and fine-grained tasks in our pipeline, we find that fewer alternatives for prompt wordings exist. Adding to the discussion on synonym set generation and the disambiguation step in \S\ref{s:limitations}, for extractive QA, a "who" question with an event name follows previous literature, and there are not many simple alternative questions for extracting the desired arguments.

\section{Monte Carlo Details}
\label{s:mcapp}

The tables containing the results discussed in Section 6 are below:
\vspace{-.5em}
\begin{table}[H]
\footnotesize
\centering
\begin{tabular}{|l|cccccc|}
\hline
\textbf{Temp} & 0&1&2&3&4&5 \\
\hline
\textbf{0} & {\color{olive}.9986} & {\color{blue}.0005} & {\color{blue}.0004} & {\color{blue}.0002} & {\color{blue}.0001} & {\color{blue}.0001}\\
\textbf{0.33} & {\color{olive}.9938} & {\color{blue}.0021} & {\color{blue}.0015} & {\color{blue}.0013} & {\color{blue}.0010} & {\color{blue}.0003}\\
\textbf{0.67} & {\color{olive}.9887} & {\color{blue}.0043} & {\color{blue}.0031} & {\color{blue}.0017} & {\color{blue}.0011} & {\color{blue}.0010}\\
\hline
\end{tabular}
\caption{The proportion of boolean answers that are different from the rest in $10$ samples over different temperatures.  }
\label{tab:bool}
\end{table}
\vspace{-.5em}

\vspace{-1em}
\begin{table}[H]
\footnotesize
\centering
\begin{tabular}{|l|cccc|}
\hline
 & 0&0.33&0.67&1 \\
\hline
\textbf{Pure Output} & {\color{olive}.9538} & {\color{blue}.8109} & {\color{blue}.6555} & {\color{blue}.5336}\\
\textbf{Aggregate substrings} & {\color{olive}.9748} & {\color{blue}.9034} & {\color{blue}.7941} & {\color{blue}.7311}\\
\hline
\end{tabular}
\caption{The proportion of extractive answers that are unanimous over $10$ samples over different temperatures. We additionally clustered answers that are substrings of each other.  }
\label{tab:ext}
\end{table}

ChatGPT results are similar, but show slightly more variability.

We also tried this procedure on Alpaca and Llama 2. Although Alpaca is able to output the same response at temperature 0 over many prompt executions unlike GPT-family models, when the temperature increases, in addition to outputting different synonym sets, it outputs the synonym sets in a different format for each execution, where the format is not easily parsable into a mathematical set representation that fits our needs. We observe similar behavior with Llama 2, where after increasing temperature, the output becomes very random and may not include English words (random letters are sometimes capitalized). Therefore, we use GPT-family models for the task but hope for future work to explore using open-source generative models, which are rapidly improving.

\section{ACE Evaluation Details}
\label{s:aceapp}

In \S\ref{s:eval}, we discussed the ACE dataset that we used to evaluate event detection and argument extraction of our method. In addition to using it to evaluate EE for its popularity, we use it because it has a variety of advantages over other datasets including whole-document annotations, realistically non-lexical-specific event classes, event modality, English data, specification of event arguments, and discourse-level entities \cite{cai-oconnor-2023-evaluating}.

\textbf{Details about input for the ACE evaluation.} Our event detection evaluation was over all $33$ event subclasses in ACE, similar to most other evaluations. For $n_t$, we used each of the $33$ subclass names. However, if a subclass in ACE is actually described as two subclasses (e.g. \texttt{arrest-jail}) with separate definitions for each, we consider the events separately (e.g. \texttt{arrest} and \texttt{jail}), aggregating their counts during evaluation. For definitions $d_t$, we used short one-sentence descriptions or paraphrases in the ACE documentation; for $k_t \in K_t$, we only add keywords if the documentation emphasizes certain words as being associated with an event class, such as `gunfire' for `attack'. 

For dyadic argument extraction, argument role pairs in ACE that map to \textit{agent} and \textit{patient} actors in \S\ref{s:task} include \textit{\{Agent, Victim\}} (e.g.~\textsc{Injure,Die}), \textit{\{Attacker,Target\}} (e.g.~\textsc{Attack}), \textit{\{Agent,Person\}} (e.g.~\textsc{Nominate, Elect, Arrest-Jail, Execute, Extradite}), \textit{\{Prosecutor,Defendant\}} (e.g.~\textsc{Trial-Hearing, Charge-Indict, Appeal}), \textit{\{Plaintiff,Defendant\}} (e.g.~\textsc{Sue}), and \textit{\{Adjudicator,Defendant\}} (e.g.~\textsc{Convict, Sentence, Acquit, Pardon}). We map paired argument roles of 20 event classes in ACE to our \textit{agent} and \textit{patient} construction.

\textbf{Terminology.} We also note a slight difference in terminology for reporting performance on ACE from some other EE evaluations: while some evaluations report trigger identification and trigger classification as separate tasks with unique F1 performance scores, our method performs the two tasks simultaneously and we refer to their combination as simply \textit{event detection}, reporting a single F1 performance score for them. Similarly, some evaluations report argument identification and argument classification separately with unique F1 scores, but our method performs the two tasks concurrently and we refer to their combination as \textit{argument extraction} in Table 2 of \S\ref{s:eval}.

\section{Affiliation Detection Details}
\label{s:affapp}

\textbf{Rebel groups.} For the affiliation detection case study (\S\ref{s:affil_detect}), we identify country references and rebel groups in each sentence. To identify rebel groups, we search the actor span to find "rebel..." or "insurgen..."; next, if our approach identifies that such an actor is affiliated with a country, we instead refer to that actor as being affiliated with the \textit{rebel group} for that country instead of the country itself.

\textbf{Identifying country references.} The step of the affiliation detection extension that identifies country references is the only step in our pipeline that does not use generative language models. We make this choice because much disagreement exists about what counts as a country, we find that different prompts could produce slightly different outputs, and we find many errors of false positives and negatives of conventionally agreed upon countries. Instead of using generative language models, we use a dictionary from TABARI. Since NYT typically also refers to the US using presidents, government positions, and government bodies, we additionally include post-Reagan US presidents, government positions including "Attorney General" and "Secretary General", and government bodies including "Congress" and "Administration" as potential country references for the United States only. If such references are incorrect, not referring to the United States, the subsequent affiliation detection step that queries if an argument is affiliated with a high-level entity tends to filter them out.

\textbf{Mapping country references to country names.} Since the dictionary from TABARI maps country references to 3-letter ISO3166 country codes (contra Correlates Of War [COW] codes\footnote{\url{https://correlatesofwar.org/cow-country-codes/}}), we use another dictionary to map\footnote{\url{https://github.com/janmarques/IsoCountries}} the country codes to country names.

\textbf{MC Details.} Since the NYT dataset was larger, we chose to perform MC over 6 boolean queries instead of over 9 as we did for the ACE evaluation to save time and cost. In \S\ref{s:mc}, we provided evidence that boolean outputs for temperature 0 were mostly unanimous, and 9 samples are probably not needed; even 3 would be reasonable. 
\end{document}